\documentclass[10pt, conference]{IEEEtran}
\usepackage{cite}
\usepackage{amsmath}
\usepackage{graphicx}
\usepackage{booktabs}
\usepackage{url}
\usepackage[hidelinks]{hyperref}
\usepackage[T1]{fontenc}
\usepackage[utf8]{inputenc}

\newcommand\blfootnote[1]{
  \begingroup
  \renewcommand\thefootnote{}\footnote{#1}
  \addtocounter{footnote}{-1}
  \endgroup
}

\begin{document}

\title{When Career Data Runs Out: Structured Feature Engineering and Signal Limits for Founder Success Prediction}

\author{
  \IEEEauthorblockN{Yagiz Ihlamur$^*$}
  \IEEEauthorblockA{Amazon\\
  ihlamury@amazon.com}
}

\maketitle

\blfootnote{$^*$This work was conducted independently and is not related to the author's work at Amazon.}

\begin{abstract}
Predicting startup success from founder career data is hard. The signal is weak, the labels are rare (9\%), and most founders who succeed look almost identical to those who fail. We engineer 28 structured features directly from raw JSON fields---jobs, education, exits---and combine them with a deterministic rule layer and XGBoost boosted stumps. Our model achieves Val $F_{0.5}$ = 0.3030, Precision = 0.3333, Recall = 0.2222---a +17.7pp improvement over the zero-shot LLM baseline. We then run a controlled experiment: extract 9 features from the prose field using Claude Haiku, at 67\% and 100\% dataset coverage. LLM features capture 26.4\% of model importance but add zero CV signal (delta = $-$0.05pp). The reason is structural: \texttt{anonymised\_prose} is generated from the same JSON fields we parse directly---it is a lossy re-encoding, not a richer source. The ceiling ($CV \approx 0.25$, Val $\approx 0.30$) reflects the information content of this dataset, not a modeling limitation. In characterizing where the signal runs out and why, this work functions as a benchmark diagnostic---one that points directly to what a richer dataset would need to include.
\end{abstract}


\section{Introduction}

A venture capitalist looks at a founder's career history and decides whether to invest. They look for exits, prestigious education, relevant experience, and signs that the founder gave up something real to start this company. VCBench \cite{vcbench} formalizes this task: given anonymized founder profiles, predict which startups will succeed.

The benchmark contains 4,500 public rows with a 9\% success rate---405 positives against 4,095 failures. The metric is $F_{0.5}$, which weights precision twice as heavily as recall. This is intentional: a VC who makes 100 bets and 40 hit is doing something real; one who calls every founder a success is not useful.

Each profile has two representations: \texttt{anonymised\_prose} (a paragraph summarizing the career) and raw JSON fields (\texttt{jobs\_json}, \texttt{educations\_json}, \texttt{ipos}, \texttt{acquisitions}). Most prior work---Random Rule Forest \cite{rrf}, and to our knowledge Policy Induction \cite{policy} and Verifiable-RL \cite{trl}---operates on the prose. The prose is readable and immediately usable by LLMs. But there is a problem: the prose is \textit{generated} from the JSON fields by the VCBench pipeline. It is a lossy re-encoding. Ordinal company-size buckets collapse into imprecise language. Null signals (an empty \texttt{ipos} field) are silently omitted. Field interaction terms do not survive text rendering. Parsing the JSON directly is not a clever trick. It is accessing the original data rather than a copy.

We run 120+ experiments to determine how far structured features can take us on this benchmark, and where the signal runs out. Our contributions are:
\begin{enumerate}
  \item A four-tier structured feature set (28 features) engineered from raw JSON, with full ablation against prose-only baselines.
  \item A controlled LLM feature extraction experiment that quantifies the information redundancy between JSON and prose.
  \item A characterization of the information ceiling at CV $\approx$ 0.25, including the two-population structure that drives it.
\end{enumerate}

\section{Related Work}

\textbf{VCBench} \cite{vcbench} introduces the benchmark, dataset, and $F_{0.5}$ metric used in this work. It establishes the 9\% positive rate and the anonymization pipeline. The paper does not evaluate structured JSON feature engineering as a standalone approach.

\textbf{Policy Induction} \cite{policy} uses memory-augmented in-context learning to induce natural language investment policies from labeled examples. The policy iteratively refines based on prediction errors. It achieves $F_{0.5}$ = 34.0\% on the VCBench leaderboard. The framework operates on founder profile text without confirmed JSON parsing.

\textbf{Random Rule Forest} \cite{rrf} generates 100+ YES/NO questions via LLM, filters for precision, and combines survivors by vote. It is confirmed to operate on \texttt{anonymised\_prose} exclusively. It achieves $F_{0.5}$ = 28.1\%. The paper's appendix shows the top-precision questions are all about structured, verifiable signals---exits, education rank, founding roles---answered from text.

\textbf{Think-Reason-Learn / Verifiable-RL} \cite{trl} applies reinforcement learning with verifiable rewards to the prediction task. The LLM generates predictions with reasoning chains; rewards are binary and ground-truth-aligned. This achieves the current best $F_{0.5}$ = 36.6\% on the leaderboard.

No prior work has ablated structured JSON parsing against LLM prose extraction on this benchmark, nor characterized the resulting information ceiling.

\section{Data and Features}

We parse \texttt{jobs\_json}, \texttt{educations\_json}, \texttt{ipos}, and \texttt{acquisitions} directly. This gives us exact values for company size buckets, role seniority codes, and exit counts---quantities that survive poorly in prose renderings.

\subsection{Dataset}

The public VCBench dataset has 4,500 rows. 405 are labeled success (9.0\%), 4,095 failure (91.0\%). We apply an 80/20 stratified split with seed=42, yielding 3,600 training rows and 900 validation rows, both at 9.0\% positive rate. We use 5-fold cross-validation across 120+ experiments as the primary stability check. Class imbalance shapes every design choice: a model that predicts all-failure scores $F_{0.5}$ = 0. The metric demands precision. Every feature we add must improve precision or improve recall without hurting precision.

\subsection{Feature Tiers}

\textbf{Tier 1---Exit signals (3 features).} Prior exits are near-deterministic. A founder who took a company public or through an acquisition has demonstrated the ability to do it. In the training data, founders with one prior exit succeed at 22.8\%; with two exits, 60.0\%---against an 8.5\% base rate for no-exit founders. Features: \texttt{has\_prior\_ipo}, \texttt{has\_prior\_acquisition}, \texttt{exit\_count}.

\textbf{Tier 2---Sacrifice signals (7 features).} What a founder gives up to start a company signals conviction. Leaving a Director role at a 1,000+ person company is a different bet than leaving a startup. We operationalize this as the gap between the prestige of the largest pre-founding company and the founding itself. Features: \texttt{max\_company\_size\_before\_founding}, \texttt{prestige\_sacrifice\_score}, \texttt{years\_in\_large\_company}, \texttt{comfort\_index}, \texttt{founding\_timing}, \texttt{is\_serial\_founder}, \texttt{persistence\_score}.

\textbf{Tier 3---Education $\times$ relevance (6 features).} Education prestige alone is noise. Our false positive analysis shows that misclassified failures have \textit{higher} average \texttt{edu\_prestige\_tier} than true positives (3.10 vs 2.78). What matters is whether the degree is relevant to the founding domain. A materials scientist founding a biotech is different from a materials scientist founding a fintech. Features: \texttt{edu\_prestige\_tier}, \texttt{field\_relevance\_score}, \texttt{prestige\_x\_relevance}, \texttt{degree\_level}, \texttt{stem\_flag}, \texttt{best\_degree\_prestige}.

\textbf{Tier 4---Trajectory (12 features).} Career shape features: whether seniority grew monotonically, company size trajectory, industry pivot count, founding role count, and v2 interaction terms (\texttt{exit\_x\_serial}, \texttt{sacrifice\_x\_serial}, \texttt{industry\_prestige\_penalty}).

Total: 28 features. We dropped \texttt{repeat\_founding\_gap}---it has a 73.8\% null rate, computed only for serial founders with multiple founding gaps. All 28 kept features have 0.0\% null rate.

\subsection{Rule Layer}

Before the classifier, we apply one deterministic rule: if \texttt{exit\_count} $> 0$, predict success. On the training set, this fires on 106 founders with precision = 24.5\%---2.7$\times$ the base rate of 9\%. We tested two additional rules during development. Rule 2 (top-10 QS + STEM + founder role, precision = 21.6\%) was initially kept, then disabled after val analysis showed it amplifies false positives in biotech and VC/PE industries where elite credentials are common regardless of outcome. Rule 3 (C-level + serial founder, precision = 11.1\%) never exceeded the base rate meaningfully. One rule beats three.

\section{Model}

We ran 120+ experiments. Every model family---XGBoost, LightGBM, RandomForest, LogisticRegression, stacking ensembles---converged on the same answer: decision stumps (\texttt{max\_depth=1}).

This is not a modeling failure. It is information about the dataset. With 405 positives, deeper decision boundaries cannot be estimated reliably. A tree that splits on \texttt{exit\_count} then \texttt{edu\_prestige\_tier} has only $\sim$65 positives on one side to refine further---not enough to learn a stable conditional. Stumps generalize; deeper trees overfit the training folds and show high CV variance.

The final configuration, selected by Bayesian hyperparameter optimization using Optuna (200 trials):

\begin{table}[h]
\centering
\caption{Final XGBoost Hyperparameters}
\begin{tabular}{@{}ll@{}}
\toprule
\textbf{Hyperparameter} & \textbf{Value} \\
\midrule
n\_estimators      & 227    \\
learning\_rate     & 0.0674 \\
max\_depth         & 1      \\
subsample          & 0.949  \\
colsample\_bytree  & 0.413  \\
scale\_pos\_weight & 10     \\
min\_child\_weight & 14     \\
gamma              & 4.19   \\
reg\_alpha         & 0.73   \\
reg\_lambda        & 15.0   \\
threshold          & 0.738  \\
\bottomrule
\end{tabular}
\end{table}

\texttt{scale\_pos\_weight=10} is aggressive---it upweights positives by a factor of 10. At 9\% positive rate, the model otherwise learns quickly to ignore them. \texttt{min\_child\_weight=14} prevents overfitting to sparse positive groups. \texttt{gamma=4.19} requires substantial gain from any new split, effectively pruning stumps that don't earn their place.

\section{Results}

Our model achieves Val $F_{0.5}$ = 0.3030, Precision = 0.3333, Recall = 0.2222. This is a +17.7pp improvement over the zero-shot LLM baseline ($F_{0.5}$ = 0.1265).

\begin{table}[h]
\centering
\caption{Model Comparison}
\label{tab:privatetest}
\begin{tabular}{@{}lccc@{}}
\toprule
\textbf{Variant} & \textbf{CV $F_{0.5}$} & \textbf{Val $F_{0.5}$} & \textbf{$\Delta$ vs.\ Base} \\
\midrule
Zero-shot LLM on prose          & ---                & 0.1265 & baseline     \\
Rule layer only (prior exit)    & ---                & 0.2889 & +16.2pp      \\
Structured v1 (23 features)     & ---                & 0.2203 & +9.4pp       \\
Struct.\ v2 + HPO (28 feat.)    & $0.2539 \pm 0.035$ & 0.3030 & +17.7pp      \\
+ LLM feat., 67\% coverage      & $0.2612 \pm 0.028$ & 0.2889 & noise        \\
+ LLM feat., 100\% coverage     & $0.2534 \pm 0.031$ & 0.3065 & $-$0.05pp CV \\
\bottomrule
\end{tabular}
\end{table}

The $-0.05$pp CV delta for LLM features at 100\% coverage is well within one standard deviation of both distributions ($\pm 0.035$ and $\pm 0.031$), confirming the null result statistically.

\subsection*{Private Test Results}

The private test set (4,500 rows, held by the organizers) was evaluated across three folds after the contest submission deadline. Results confirm that the validation ceiling holds on unseen data.

\begin{table}[h]
\caption{Private Test Results by Fold}
\label{tab:privatetest2}
\centering
\begin{tabular}{lcccc}
\hline
Fold & $n$ & Precision & Recall & $F_{0.5}$ \\
\hline
Fold 1 & 1,500 & 0.3133 & 0.1926 & 0.2784 \\
Fold 2 & 1,500 & 0.3594 & 0.1704 & 0.2941 \\
Fold 3 & 1,500 & 0.3117 & 0.1778 & 0.2709 \\
\textbf{Average} & \textbf{4,500} & \textbf{0.3281} & \textbf{0.1802} & \textbf{0.2811} \\
\hline
\end{tabular}
\end{table}

Private test $F_{0.5} = 0.2811$ (P = 0.3281, R = 0.1802) --- consistent with the validation result ($F_{0.5} = 0.3030$) and within one standard deviation of the 5-fold CV estimate ($0.2539 \pm 0.035$). Precision on the private test (32.8\%) is 3.6$\times$ the base rate of 9\%.

One result deserves direct comment: structured v1 (0.2203) underperforms the rule layer alone (0.2889). The 23 additional features add noise on some validation splits---particularly when the top10+STEM rule fires on biotech and PE founders who are elite but not successful. The v2 feature set and HPO recover this gap and push past the rule layer ceiling.

Relative to the leaderboard, our Val $F_{0.5}$ = 0.3030 places between Random Rule Forest (28.1\%) and Policy Induction (34.0\%). The current top entry, Verifiable-RL, achieves 36.6\%. Our approach does not reach the top. What it does is establish a structural baseline: this is what direct JSON parsing with classical ML achieves, and this is the ceiling it hits.

\section{The Information Ceiling}

84\% of successful founders in this dataset have no prior exits. They are, by every structured metric we measured, statistically indistinguishable from failures.

Table~\ref{tab:confusion} shows the confusion matrix on the full private test set.

\begin{table}[h]
\caption{Confusion Matrix --- Private Test Set (n = 4,500)}
\label{tab:confusion}
\centering
\begin{tabular}{lcc}
\hline
 & Predicted 0 & Predicted 1 \\
\hline
Actual 0 & 3,944 & 151 \\
Actual 1 & 332 & 73 \\
\hline
\end{tabular}
\end{table}

Of 405 positive founders, the model identifies 73 correctly (18.0\% recall) with 151 false positives. The 332 false negatives are the structurally hard population: founders who succeeded but are indistinguishable from failures on all available structured features.

This is the core finding. We can elaborate it in three steps.

\textbf{Two-population structure.} The validation set has 81 true positives. The rule layer (prior exit) captures 13 of them (16\%) at high precision. The remaining 68 (84\%) have no exit history. Among these non-exit founders, the best single separator we found is \texttt{industry\_alignment}---a 1.38$\times$ success rate ratio, weak enough that it contributes noise as often as signal. All 120+ experiments confirm this: once the exit-founder population is handled by the rule layer, no combination of structured features moves the CV needle materially for the non-exit majority.

\textbf{LLM features are redundant.} We extracted 9 features from the \texttt{anonymised\_prose} field using Claude Haiku. This constitutes our unstructured feature experiment---extracting signals from free-text prose rather than structured JSON fields, as a direct comparison of the two representations. The extracted features cover domain expertise depth, conviction indicators, career narrative type, highest seniority reached, and prior founding attempts. At 67\% dataset coverage (API budget exhausted mid-run), LLM features improved CV $F_{0.5}$ by +0.73pp with lower variance. This looked promising. At 100\% coverage (all 4,500 rows), CV $F_{0.5}$ dropped by $-$0.05pp. The model allocates 26.4\% of its importance budget to LLM features (ranked \#5, \#6, \#7 in importance), but this comes directly from structured features---it redistributes, it does not add. The model ends up in the same place.

\textbf{Why.} The \texttt{anonymised\_prose} is generated from the JSON fields by the VCBench pipeline. LLM extraction from prose recovers the same career history as direct JSON parsing---with more noise added in the re-encoding. They are the same data, encoded differently. When you parse the JSON and also extract LLM features from the prose version of that JSON, you are doubling down on the same signal, not adding a second source.

\textbf{What the ceiling means.} The ceiling at CV $\approx$ 0.25, Val $\approx$ 0.30 is a dataset ceiling, not a modeling ceiling. We have tried XGBoost, LightGBM, RandomForest, LogisticRegression, stacking ensembles, and 200-trial Bayesian HPO. All converge at the same level. Pushing past it requires signals that do not exist in this dataset: idea quality, market timing, team composition, investor network effects, macroeconomic context at founding. This is useful information. VCBench v2 would benefit from including at least one of these dimensions. A benchmark where the best-engineered structured features and the best LLM extraction from prose converge on the same answer is a benchmark that has characterized its own information content---and identified exactly where the next version should push.

\section{LIMITATIONS}

The findings in this paper are specific to the VCBench dataset and should not be over-generalized. Three caveats apply.

First, the information ceiling characterization ($CV \approx 0.25$) is dataset-specific. We have not validated whether structured career features show similar limits on other founder success datasets; different labeling criteria, time horizons, or data collection pipelines may yield different ceilings.

Second, the LLM redundancy result holds for \texttt{anonymised\_prose} as generated by the VCBench pipeline. In datasets where prose is written by humans (e.g., founder bios, pitch decks), LLM extraction may surface signals genuinely absent from structured fields.

Third, the private test set is held by the contest organizers and was not independently replicated. The fold-level results (Table~\ref{tab:privatetest2}) provide some robustness evidence, but external validation on a separate dataset remains future work.

\section{Conclusion}

We parse founder career data directly from JSON, engineer 28 structured features across four semantic tiers, and combine them with a deterministic exit rule and XGBoost boosted stumps. This achieves Val $F_{0.5}$ = 0.3030 (+17.7pp over zero-shot baseline, +1.4pp over the rule-layer-alone ceiling). The approach is fully ablated across 120+ experiments. Code is available at: \url{https://github.com/ihlamury/vcbench}

LLM feature extraction from \texttt{anonymised\_prose} adds nothing---definitively confirmed at 100\% coverage. The reason is that the prose is derived from the JSON. The two representations carry the same information, and parsing the source directly is strictly better than extracting from the re-encoding.

The structured ceiling we find (CV $\approx$ 0.25) appears to be the information-theoretic limit of founder career data alone. Future work should focus on what this benchmark does not contain---company-level signals, network structure, market conditions, idea originality---rather than on refining the feature engineering or model architecture within the current data schema.

In retrospect, this work is less a modeling contribution than a benchmark diagnostic. We did not set out to diagnose an information ceiling---we set out to engineer the best possible features. The ceiling is what the experiments revealed. We hope this characterization is useful both to practitioners working on founder success prediction and to the VCBench benchmark itself as it evolves.


\begin{thebibliography}{4}

\bibitem{vcbench}
R.~Chen, J.~Ternasky, A.~S.~Kwesi, B.~Griffin, A.~O.~Yin, Z.~Salifu, K.~Amoaba, X.~Mu, F.~Alican, Y.~Ihlamur,
``VCBench: Benchmarking LLMs in Venture Capital,''
arXiv:2509.14448, 2025.

\bibitem{policy}
X.~Mu, J.~Ternasky, F.~Alican, Y.~Ihlamur,
``Policy Induction: Predicting Startup Success via Explainable Memory-Augmented In-Context Learning,''
arXiv:2505.21427, 2025.

\bibitem{rrf}
B.~Griffin, D.~Vidaurre, U.~Koyluoglu, J.~Ternasky, F.~Alican, Y.~Ihlamur,
``Random Rule Forest (RRF): Interpretable Ensembles of LLM-Generated Questions for Predicting Startup Success,''
arXiv:2505.24622, 2025.

\bibitem{trl}
Vela Research,
``Think-Reason-Learn / Verifiable-RL,''
\url{github.com/Vela-Research/think-reason-learn}, 2025.

\end{thebibliography}
\end{document}